# MedPerf: Open Benchmarking Platform for Medical Artificial Intelligence using Federated Evaluation

Alexandros Karargyris* (IHU Strasbourg), Renato Umeton* (Dana-Farber Cancer Institute; Weill Cornell Medicine; Harvard T.H. Chan School of Public Health; Massachusetts Institute of Technology), Micah J. Sheller* (Intel), Alejandro Aristizabal (Factored AI), Johnu George (Nutanix), Srini Bala (Supermicro), Daniel J. Beutel (University of Cambridge; Adap), Victor Bittorf (Facebook), Akshay Chaudhari (Stanford University School of Medicine; Stanford University), Alexander Chowdhury (Dana-Farber Cancer Institute), Cody Coleman (Stanford University), Bala Desinghu (Rutgers University), Gregory Diamos (Landing.AI), Debo Dutta (Nutanix), Diane Feddema (Red Hat), Grigori Fursin (OctoML), Junyi Guo (Harvard T.H. Chan School of Public Health), Xinyuan Huang (Cisco), David Kanter (MLCommons), Satyananda Kashyap (IBM Research), Nicholas Lane (University of Cambridge; Samsung), Indranil Mallick (Tata Medical Center), Pietro Mascagni (Fondazione Policlinico Universitario A. Gemelli IRCCS; IHU Strasbourg), Virendra Mehta (University of Trento), Vivek Natarajan (Google), Nikola Nikolov (Supermicro), Nicolas Padoy (University of Strasbourg), Gennady Pekhimenko (University of Toronto; Vector Institute), Vijay Janapa Reddi (Harvard University), G Anthony Reina (contributed while at Intel), Pablo Ribalta (NVIDIA), Jacob Rosenthal (Dana-Farber Cancer Institute; Weill Cornell Medical), Abhishek Singh (Massachusetts Institute of Technology), Jayaraman J. Thiagarajan (Lawrence Livermore National Laboratory), Anna Wuest (Harvard T.H. Chan School of Public Health), Maria Xenochristou (Stanford University), Daguang Xu (NVIDIA), Poonam Yadav (York University), Michael Rosenthal (Brigham And Women's Hospital; Dana-Farber Cancer Institute; Harvard Medical School), Massimo Loda (Weill Cornell Medicine; Dana-Farber Cancer Institute; Broad Institute of MIT and Harvard), Jason M. Johnson (Dana-Farber Cancer Institute), Peter Mattson (Google; MLCommons)

https://mlcommons.org/en/#founders

* These authors contributed equally. Correspondence to be addressed to:
alexandros.karargyris@ihu-strasbourg.eu, renato_umeton@dfci.harvard.edu, micah.j.sheller@intel.com, jason_johnson@dfci.harvard.edu, peter@mlcommons.org.

# Abstract

Medical AI has tremendous potential to advance healthcare by supporting the evidence-based practice of medicine, personalizing patient treatment, reducing costs, and improving provider and patient experience. We argue that unlocking this potential requires a systematic way to measure the performance of medical AI models on large-scale heterogeneous data. To meet this need, we are building MedPerf, an open framework for benchmarking machine learning in the medical domain. MedPerf will enable federated evaluation in which models are securely distributed to different facilities for evaluation, thereby empowering healthcare organizations to assess and verify the performance of AI models in an efficient and human-supervised process, while prioritizing privacy. We describe the current challenges healthcare and AI communities face, the need for an open platform, the design philosophy of MedPerf, its current implementation status, and our roadmap. We call for researchers and organizations to join us in creating the MedPerf open benchmarking platform.

**Code availability**: we made all code available under an Apache license at https://github.com/mlcommons

# 1 Introduction - Need for Wide Data Access and Model Generalization

As medical AI has begun to transition from research to clinical care[1–4], national agencies around the world have started drafting regulatory frameworks to support this new class of interventions. Examples include the US Food and Drug Administration (https://www.fda.gov/medical-devices/digital-health-center-excellence), the European Medicines Agency (https://www.ema.europa.eu/en/about-us/how-we-work/regulatory-science-strategy), and the Central Drugs Standard Control Organisation in India[5]. A key point of agreement for all regulatory agency efforts is a requirement for formal, large-scale validation of medical AI models[6–8]. Widespread approval and adoption of medical AI models will thus require expansion and diversification of clinical data sourced from multiple organizations. Furthermore, there are emerging parallels between stages for approval for medical AI interventions and the regulatory approval of small molecules or medical devices through clinical trials[9–11].

Pioneering research in the medical field and elsewhere[12,13] has demonstrated that using large and diverse datasets during model training results in more accurate models. Such models are also expected to be more generalizable to other clinical settings. Other studies have shown that models trained with data from limited and specific clinical settings demonstrate bias toward specific patient populations[14–16], and such data biases can lead to models that appear promising during development but have lower performance in wider deployment[17,18]. A given static model may be susceptible to distribution shifts for the model's input or the model's target, or both[19]. For example, input distributional shifts may occur when an algorithm is evaluated on a population different than the one upon which it was trained on, when there are changes to local demographics or disease prevalence, or as a result of software or hardware upgrades of medical imaging equipment used for data acquisition. Similarly, distributional shifts may also arise from variations in geographic insurance reimbursement and medical procedure trends, or from new annotation or labeling guidelines. These issues, which are often intertwined with one another and frequently result in performance degradation, can also hinder trust and acceptance of AI among healthcare stakeholders, including clinicians, patients, insurers, and regulators.

We believe a new approach to leveraging diverse data can deliver consistent clinical and business value to healthcare data owners, while creating adoption incentives through lower implementation cost and lower deployment risk[6]. Such an approach should allow collaborative model training and evaluation on large, multi-institutional and representative datasets while complying with privacy and regulatory requirements. However, the degree to which these requirements can be met during collaborative training is still an open research question[43].

Here we present MedPerf, an approach focused on broader data access during model evaluation, which we believe will best support model generalization as well as improve clinician and patient confidence. MedPerf was built upon the group's experience leading and disseminating efforts such as (i) the development of standardized benchmarking platforms (e.g., MLPerf for benchmarking machine learning training[20] and inference[21] across industries in a pre-competitive space - https://mlcommons.org/#MLPerf); (ii) the implementation of federated learning software frameworks (e.g., NVIDIA CLARA, Intel OpenFL[22], and Flower by Adap/University of Cambridge); (iii) the ideation and coordination of federated medical challenges across dozens of clinical sites and research institutes (e.g., BraTS[23] and FeTS[24]); as well as (iv) other prominent medical AI and machine learning efforts spanning multiple countries and healthcare specialties (e.g., oncology[25] and COVID[26]). MedPerf should also illuminate cases where better models are needed, increase adoption of existing generalizable models, and incentivise further model development, data annotation, curation, and data access, while preserving patient privacy. The development of this approach requires (a) consistent and rigorous methodologies to evaluate performance of AI models for real-world use in a standardized manner, (b) a technical approach that enables measuring model generalizability across institutions, while maintaining data privacy and respecting model intellectual property, and (c) a community of expert groups to employ the evaluation methodology and the technical approach to define and operate mature performance benchmarks.

MedPerf aims to address these goals. MedPerf is an open-source framework designed to develop and support benchmark reference implementations, respect data privacy, and support model evaluation through formal generation of benchmarking working groups. MedPerf provides the opportunity to set standards, best practices, and benchmarking for medical AI in a pre-competitive space. The current list of contributors includes representatives of 18 companies, 13 universities, 6 hospitals, and 10 countries.

# 2 Challenges: Risk, Cost and Uncertain Return

In this section, we discuss challenges to wider data access for AI training and evaluation in healthcare. Convincing data owners to broaden data access is hindered by substantial regulatory, legal, and public perception risks, high up-front costs, and uncertain financial return on investment.

## 2.1 Risk

Sharing patient data presents three major classes of risk: liability, regulatory, and public perception. Sharing patient data can expose providers to liability risk in multiple ways. Shared data could be stolen or misused in a manner damaging to patients (e.g., to discriminate against patients with certain conditions). Patient data are protected by complex regulations such as HIPAA in the United States and GDPR in Europe that carry substantial penalties for violators. The perception of risk is also heightened because AI is a relatively new paradigm where application of existing regulations can be unclear. Lastly, even if data are shared legally and used beneficially, people naturally value privacy, and sharing data without explicit consent could lead to negative public perception[27].

## 2.2 Cost

Sharing data requires up-front investment to turn raw data into a useful resource for AI. This transformation involves multiple steps:

1. Data collection: Cohorts need to be identified and the corresponding data need to be made accessible.
2. Transformation: Once accessible, data must be reformatted to a standardized representation for each data type (e.g., DICOM[28] for medical images) suitable for subsequent steps.
3. Anonymization: Data are anonymized by removing identifying information and/or filtering to comply with statistical and regulatory requirements (e.g., K anonymity[29]).
4. Labeling: For many AI tasks, data must be labeled (i.e., annotated) according to the task (e.g., brain tumor segmentation). To ensure quality and performance, labeling should be consistent across institutions. This step is expensive, highly human-dependent, and error-prone, while carrying additional costs related to annotation correction, versioning, and dataset maintenance[30].

5. Review: Data need to be reviewed for regulatory, legal, and policy compliance, and patients or patient groups need to be consulted for the design and perception of the use case.
6. Licensing: Data must be licensed in a manner that fulfills business and/or scientific interests while complying with existing regulations.
7. Sharing: Data must be physically shared with licensees, through complex legal agreements, which may require secure transmission of large data volumes or the creation of specially designed data enclaves.

Navigating these steps can be costly. The technical part of the process is also complex, requiring a mix of medical, artificial intelligence, and software engineering skills. There are multiple opportunities for error that may not be revealed until downstream consequences emerge, necessitating careful validation at each step, sometimes with multiple iterations[31].

## 2.3 Uncertain Return

Even if a data owner (e.g., a hospital) is willing to pay for these costs and mitigate these risks, benefits can be unclear for financial or technical reasons. For example, if the development of an AI-based solution is driven by the AI model builder instead of the data provider, the AI provider may see a greater share of the eventual benefits than the data owner, even though the data owner may incur a greater share of the risk.

From a technical perspective, it can be difficult to prove a model's performance prior to deployment. Current medical AI community challenge efforts (e.g., FeTS[24], CheXpert[32], BraTS[33], NLST[34], CHAOS[35], fastMRI[36]) have been invaluable for advancing research but lack the scope to serve as real-world evaluation mechanisms in clinical settings. These challenges typically focus on a single dataset and task and thus do not reflect the diversity (e.g., multi-modal and multi-institutional) and complexity (e.g., different clinical and technical workflows) of real-world use cases. Model training and evaluation on non-diverse datasets carries increased risk of overfitting and the chance that even top-performing models will not generalize in real-world use cases, where clinical data reside in multi-institutional, geographically distributed organizations with significant differences across domains (i.e., domain shifts)[14].

# 3 Proposed Solution: An Open Benchmarking Platform using Federated Evaluation

Our goal is to increase the clinical impact of AI by leveraging more data across multiple facilities to address the challenges described above. We are developing an *open benchmarking platform* that combines a lower-risk, evaluation-focused approach without data sharing along with appropriate infrastructure, technical support and organizational coordination. This approach can reduce the risk and cost associated with data sharing while increasing the likelihood of business and medical benefits provided by AI solutions. MedPerf should lead to wider adoption, more efficacious and cost-effective clinical practice, and improve patient outcomes.

Our technical approach uses *federated evaluation*, a reduced-risk form of federated learning. At its core, the aims of federated evaluation are to make sharing models with multiple data owners easy and reliable, to evaluate those models against data owners' data in controlled settings, and to aggregate and analyze evaluation metrics. Importantly, by limiting the goal to model evaluation, and by aggregating only evaluation metrics, federated evaluation poses significantly lower risk to patient privacy than collaborative model training, while also minimizing the risk[37,38] of intellectual property theft and data misuse.

More specifically, our open platform for federated evaluation will provide a common, open-source infrastructure for defining medical AI benchmarks and coordinating federated evaluation of models against such benchmarks. We are building the infrastructure with best practices to help align AI model owner/developers with data owners, through an active community with a neutral organization at its core. We intend for our approach to be compatible with, and to build upon, existing federated learning frameworks, rather than to compete with them. Furthermore, as detailed below, we introduce steps that give data owners control over what algorithms run on their data and allow them to confirm benchmarking results.

## 3.1 Risks are Mitigated by Focusing on Model Evaluation and Trusted Groups

MedPerf addresses regulatory, liability, and public perception risks using a three-pronged approach.

First, because the initial focus is on model evaluation instead of training, our federated evaluation approach maximizes value without data leaving the possession of data owners, either directly or accidentally leaked through results. We only need data owners to share agreed-upon evaluation metrics (e.g., specificity), which are aggregated across participating institutions before disseminating. This mitigates most regulatory, public perception, and legal risk.

Second, Medperf retains human evaluators[39] as a critical part of the proces: the MedPerf client software requires a data-owner's system administrator to approve all model evaluations and result uploads, and automatically records transactions to support auditing. Moreover, to protect against malicious or erroneous implementations, MedPerf requires that (a) all novel code has no network access and restricted local file-system access, (b) evaluation algorithm implementations are well-vetteed and common among benchmarks, and (c) all output (i.e., statistics) must be explicitly approved by data owners before it is uploaded to the MedPerf platform.

Third, we leverage social trust: we enable benchmarks to be specified, developed, and deployed publicly or within closed groups, such as provider networks with pre-existing trusted relationships and business and legal contracts, and these closed-group benchmarks will be prioritized during the pilot phase of deployment. We are developing the MedPerf infrastructure through MLCommons, a non-profit with diverse membership and open-source practices, backed by dozens of high-profile companies and institutions (https://mlcommons.org/en/#founders).

## 3.2 Costs are Reduced through Open Infrastructure and Best Practices

We aim to reduce the costs of data sharing by developing open-source infrastructure and best practices that enable infrastructure vendors, model owners, and data owners to collaboratively build within a fast-growing ecosystem.

First, we provide community best practices for sharing models and data. For instance, we are using the MLCube container for model sharing (see (https://mlcommons.org/en/mlcube/) for concept introduction and practical examples, and (https://github.com/mlcommons/mlcube) for the code repository). MLCube extends common container standards, such as Docker and

Singularity, to offer a simple and consistent file system-based interface for other infrastructure to train or make inferences using AI models (e.g., for testing harnesses or federated learning). Additionally, deployment tools like Docker and Singularity enable hospital information technology groups to evaluate the AI model code for security concerns using common methods and tools.

Second, we are developing an open-source hub for medical AI benchmarks and a consistent methodology for benchmarking. The hub will offer coordination among benchmark groups, model developers, and data owners by providing a central model and data registry and by storing results, but will not directly handle proprietary models or data, ensuring that these assets remain in the hands of their owners. Instead, model and data owners will register hashes to enable checking the integrity of their assets without exposing them to the platform. This method will ensure that benchmark results can be compared to better establish promising technical approaches.

## 3.3 Return on Investment: Increasing Certainty through Better Model Evaluation

Our approach decreases the uncertainty of deploying AI models by enabling easy evaluation against data held by multiple data owners. We enable model developers to indirectly interact with data owners' datasets and thus tap into a large, *virtual test set*. In doing so, we increase the size of the test set and thereby reduce uncertainty of the evaluation - even if all the data are from a single provider. More importantly, by enabling evaluation against data from multiple providers, we can more effectively evaluate how the model will perform when deployed at different facilities with diverse patient populations. And by providing multi-site performance feedback to model developers, we increase the odds of successful model deployment. Ultimately, demonstration that broad evaluation via federated evaluation is correlated with clinical efficacy will further improve clinician and patient confidence and motivate additional data owners to participate.

## 3.4 Building an On-Ramp to Federated Learning

We believe widespread adoption of federated evaluation will also spur wider adoption of federated learning. Federated learning (FL) is a promising technology to enable development of

AI models by leveraging data from multiple institutions without directly sharing data[40–42]. While FL enables model training without data sharing, data may leak through the model parameters themselves, requiring additional mitigations[43–45]. Research and development of these mitigations is ongoing, slowing the adoption of the technology. We believe that federated evaluation provides concrete benefits while building industry familiarity with the technology needed for full FL.

# 4 MedPerf Technical Approach

In this section, we describe the structure and functionality of an open benchmarking platform for medical AI. We define a MedPerf benchmark in this context, discuss user roles required to successfully operate a benchmark, and provide an overview of the operating workflow.

## 4.1 What is a Benchmark

For the purposes of our platform, a benchmark is a bundle of assets that enables quantitative measurement of the performance of AI models for a specific clinical problem. A benchmark consists of the following major components:

1. **Specifications**: precise definition of the clinical setting (e.g., problem or task and specific patient population) on which trained AI models are to be evaluated, the labelling methodology, and specific evaluation metrics.
2. **Dataset Preparation**: a process that prepares datasets for use in evaluation, and can also test the prepared datasets for quality and compatibility.
3. **Registered Datasets**: a list of registered datasets prepared according to the benchmark criteria and approved for evaluation use by their owners, e.g. patient data from multiple facilities representing (as a whole) a diverse patient population.
4. **Evaluation:** a consistent implementation of the testing pipelines and evaluation metrics.
5. **Reference Implementation:** an example of a benchmark submission consisting of example model code, the evaluation component above, and publicly available de-identified or synthetic sample data.
6. **Registered Models**: a list of registered models to run in this benchmark.
7. **Documentation:** documents for understanding and using the benchmark.

Our platform uses the MLCube container for components such as Dataset Preparation, Evaluation, and the Registered Models. MLCube containers are software containers (e.g., Docker and Singularity) with standard metadata and a consistent file-system level interface. By using MLCube, the infrastructure software can easily interact with models implemented using different approaches and/or frameworks, running on different hardware platforms, as well as leverage common software tools for validating proper secure implementation practices (e.g., CIS Docker Benchmarks).

## 4.2 Benchmarking User Roles

We have identified four primary roles in operating an open benchmark platform, outlined in Table 1. In many cases, a single organization may participate in multiple roles, and multiple organizations may share any given role. Beyond these roles, the long term success of medical AI benchmarking requires organizations that create and adopt appropriate community standards for interoperability such as Vendor Neutral Archives (VNA)[46,47], DICOM[28], OMOP[48] (https://www.ohdsi.org/data-standardization/the-common-data-model/), PRISSMM[49], and HL7/FHIR[50].

## 4.3 Benchmarking Workflow

Our open benchmarking platform uses the workflow depicted in Figure 1. To start, a benchmark group registers the benchmark with the benchmarking platform (1) and then recruits data (2) and model owners (3). The benchmarking platform sends model evaluation requests to the data owners who approve and execute the evaluations, successively vetting and then pushing results to the benchmarking platform (4). The benchmarking platform shares the results with participants based on a policy specified by the benchmark group (5). Table 2 provides further details about each workflow step.

# 5 MedPerf Roadmap

Ultimately, we aim to deliver an open platform that enables groups of researchers and developers to use federated evaluation to provide high-confidence evidence of generalized model performance to regulators, health care providers, and patients. In Table 3, we review necessary steps, scope of each step, and current progress towards developing this open benchmarking platform.

## 6 Related Work

Our effort is inspired by several classes of related work. First, we adopt a federated approach to data, focusing first on evaluation to lower the barriers to adoption. Second, we adopt the standardized measurement approach to medical AI from organizations such as RSNA (https://www.rsna.org), SIIM (https://siim.org), and Kaggle (https://www.kaggle.com), and we generalize these efforts to a standard platform that can be applied to many problems rather than focus on a specific one. Third, we leverage the open, community-driven approach to benchmark development successfully employed to accelerate hardware development, through efforts such as MLPerf (https://mlcommons.org) and SPEC (https://www.spec.org/benchmarks.html), and apply it to the medical domain. Lastly, we push towards creating shared best practices for AI as inspired by efforts like MLflow (https://mlflow.org) and Kubeflow (https://www.kubeflow.org) for AI operations, as well as MONAI (https://monai.io) and GaNDLF (https://cbica.github.io/GaNDLF/) for medical models.

## 7 Discussion and Conclusion

Our initial goal is to provide medical AI researchers with reproducible benchmarks based on diverse patient populations to assist healthcare algorithm development. We believe such benchmarks will increase development interest and solution quality, leading to patient benefit and growing adoption. Furthermore, our platform will help advance research related to, but not limited to, data utility, robustness to noisy annotations, and understanding of model failures. If a critical mass of AI researchers adopts these benchmarks, healthcare decision makers will see substantial benefits from aligning with this effort to increase benefit for their patient populations. Ultimately, standardizing best practices and evaluation methods will lead to highly accurate models that are acceptable to regulatory agencies and clinical experts, and create momentum within patient advocacy groups. By bringing together these diverse groups, starting with AI researchers and healthcare organizations, as well as building trust with clinicians, regulatory authorities, and patient advocacy groups, we envision accelerating the adoption of AI in healthcare and increased clinical benefits to patients and providers worldwide.

However, we cannot achieve these benefits without the help of the technical and medical community. We call for:
- Healthcare stakeholders to form the benchmark groups that define benchmark specifications and oversee the analyses of their results.
- AI researchers to test this end-to-end platform and use it to create and validate their own models across multiple institutions.
- Data owners (e.g., healthcare organizations) to register their data in the platform, again while never sharing the data itself.
- Data model standardization efforts to enable collaboration between institutions, such as the OMOP Common Data Model and VNA.
- Regulatory bodies to develop medical AI solution approval requirements that include technically robust and standardized benchmarking.

We believe open efforts like MedPerf can drive innovation and bridge the gap between AI research and real-world clinical impact. To achieve these benefits, there is a critical need for broad collaboration, reproducible, standardized and open computation, and a passionate community that spans academia, industry, and the clinical world. With MedPerf, we aspire to bring such a community of stakeholders together as a critical step toward realizing the grand potential of medical AI. We invite participation at [https://mlcommons.org/medperf](https://mlcommons.org/medperf).

# Figures and Tables

## Figures

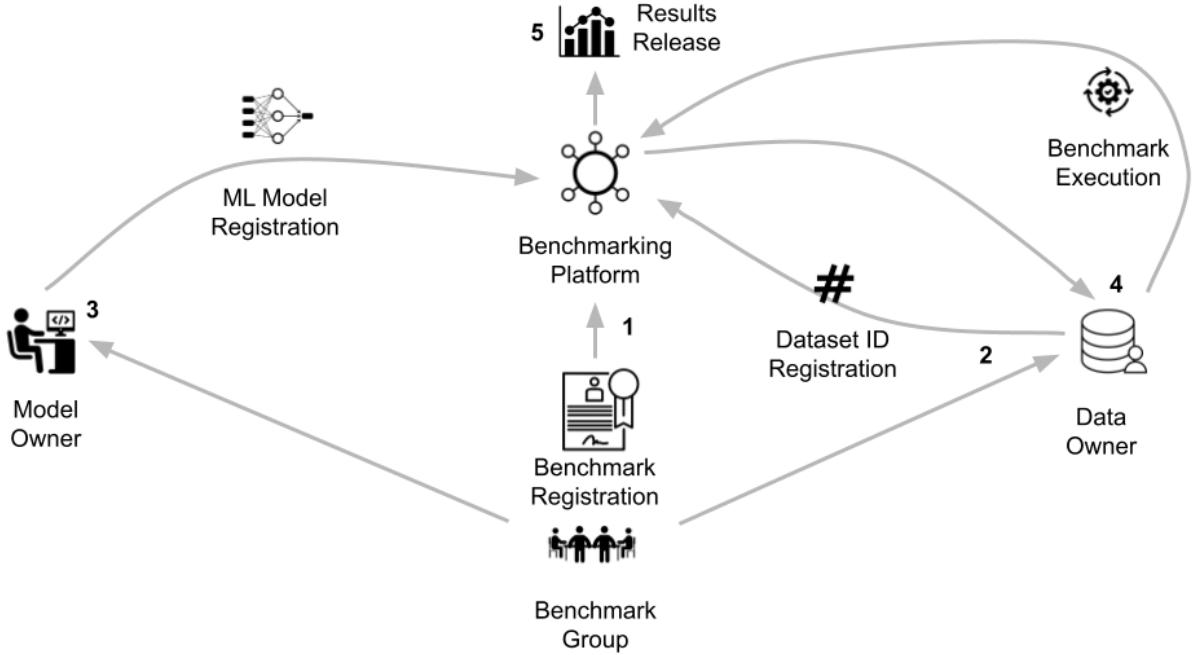

**Figure 1.** Benchmarking workflow, from benchmark registration to results. See Table 2 for details of all workflow steps, 1 through 5.

# Tables

| Table 1. Benchmarking user roles and responsibilities. | | |
|---|---|---|
| Role name | Role definition | Role responsibilities |
| Benchmark group | Benchmark groups include regulatory bodies, groups of experts (e.g., clinicians, patient representative groups), and data or model owners wishing to drive evaluation of their model or data | ● Authors the benchmark, manages all benchmark assets, and may produce some assets (e.g., dataset preparation)<br>● Recruits model owners and data owners, makes an open benchmark for model owners, and approves applicants<br>● Controls access to the aggregated statistical results |
| Data owner | Data owners may include hospitals, medical practices, research organizations, and healthcare insurance providers that own medical data, register medical data, and execute benchmark requests | ● Registers data with benchmarking platform<br>● Performs data labelling<br>● Downloads and executes a data preparation processor to prepare data<br>● Downloads and periodically uses platform client to approve and serve requests and to approve and upload results to or from benchmarking platform |
| Model owner | Model owners include AI researchers and software vendors that own a trained medical AI model and want to evaluate its performance. | ● Registers model with benchmarking platform<br>● Views results of their model on the benchmark<br>● Has the option to approve sharing of results of that benchmark with other model/data owners or the public if allowed by benchmark group |
| Platform provider | Organizations like MLCommons that operate a platform that enables benchmark groups to run benchmarks by connecting data owners with model owners | ● Manages user accounts and provides a website for registering and discovering benchmarks, datasets, models, and for the overall workflow management<br>● Coordinates active benchmarks by sending requests, aggregating results, and managing result access |

**Table 2.** Benchmarking workflow, steps, and interconnections with roles.

| Workflow step | Objective |
|---|---|
| **1** Define and register benchmark | ● The benchmarking process starts with establishing a benchmark group of healthcare stakeholders: healthcare organizations, clinical experts, AI researchers, and patient advocacy groups<br>● Benchmark group identifies a clinical problem where an effective AI-based solution can have significant clinical impact<br>● Benchmark group registers the benchmark on the platform; and provides the benchmark assets (see "What is a Benchmark") |
| **2** Recruit data owners | ● Benchmark group recruits data and model owners either by inviting trusted parties or by making an open call for participation<br>● Dataset owners are recruited to maximize aggregate dataset size and diversity on a global scale. Many benchmarking efforts may initially focus on data providers with existing agreements |
| Prepare and register data sets | ● In coordination with the benchmark group, dataset owners are responsible for data preparation (i.e., extraction, preprocessing, labelling, reviewing for legal/ethical compliance)<br>● Once the data is prepared and approved by the data owner, the dataset can be registered with the benchmarking platform |
| **3** Recruit model owners | ● Model owners modify the benchmark reference implementation. To enable consistent execution on data owner systems, the solutions are packaged inside MLCube containers<br>● Model owners must conduct appropriate legal and ethical review prior to submission of a solution for evaluation |
| Prepare and register models | ● Once implemented by the model owner and approved by the benchmark group, the model can be registered on the platform |
| **4** Execute benchmarks | ● Once the benchmark, dataset, and models are registered to the benchmarking platform, the platform notifies the data owners that models are available for benchmarking<br>● The data owner runs a benchmarking client that downloads available models, reviews and approves models for safety, then approves execution<br>● Once execution completes, the data owner reviews and approves upload of the results to the benchmark platform |
| **5** Release results | ● Benchmark results are aggregated by the benchmarking platform and shared per the policy specified by the benchmark group |

**Table 3.** MedPerf roadmap stages, scopes, and corresponding details for each stage.

| Roadmap stage | Scope - Status | Details |
| --- | --- | --- |
| Design | An open benchmarking platform called MedPerf - COMPLETE | MedPerf is supported by the non-profit MLCommons Association. MLCommons brings together engineers and academics globally to make AI better for all; they have already created and host the MLPerf benchmark suites for AI performance (as measured by speed-up, electrical consumption, and other metrics) |
| Implementation of sample benchmarks to drive development | Computed tomography (CT) of the abdomen - IN PROGRESS<br><br>Chest x-ray - IN PROGRESS<br><br>Brain tumor segmentation - IN PROGRESS | We chose these motivating problems because they (1) affect a large, global patient population and represent a substantial opportunity for clinical impact, (2) have high-potential AI solutions and (3), have public datasets and open-source models in development |
| Implementation of platform | Phase 1: Single-system proof-of-concept - IN PROGRESS | Demonstrate technical approach using public data and open-source models on a single system that simulates multiple systems (which eliminates platform incompatibility and communication issues) |
| | Phase 2: Distributed proof-of-concept - PLANNED | Demonstrate technical approach using public data and open source models communicating across the internet on multiple systems belonging to potential data and model owners |
| Deployment of platform | Phase 1: Early deployment - PLANNED | Launch 2-3 benchmarks targeting high-impact medical problems |
| | Phase 2: Wide-scale deployment - PLANNED | Multiple benchmarking efforts |

# References

1. Benjamens, S., Dhunnoo, P. & Meskó, B. The state of artificial intelligence-based FDA-approved medical devices and algorithms: an online database. *npj Digital Med.* **3**, 118 (2020).

2. Wang, P. *et al.* Effect of a deep-learning computer-aided detection system on adenoma detection during colonoscopy (CADe-DB trial): a double-blind randomised study. *Lancet Gastroenterol. Hepatol.* **5**, 343–351 (2020).

3. Yao, X. *et al.* Artificial intelligence-enabled electrocardiograms for identification of patients with low ejection fraction: a pragmatic, randomized clinical trial. *Nat. Med.* **27**, 815–819 (2021).

4. Beede, E. *et al.* A Human-Centered Evaluation of a Deep Learning System Deployed in Clinics for the Detection of Diabetic Retinopathy. in *Proceedings of the 2020 CHI Conference on Human Factors in Computing Systems* 1–12 (ACM, 2020). doi:10.1145/3313831.3376718.

5. Verma, A., Rao, K., Eluri, V. & Sharm, Y. *Regulating AI in Public Health: Systems Challenges and Perspectives*. (2020).

6. Wu, E. *et al.* How medical AI devices are evaluated: limitations and recommendations from an analysis of FDA approvals. *Nat. Med.* **27**, 582–584 (2021).

7. Vokinger, K. N., Feuerriegel, S. & Kesselheim, A. S. Continual learning in medical devices: FDA's action plan and beyond. *Lancet Digit. Health* **3**, e337–e338 (2021).

8. Kann, B. H., Hosny, A. & Aerts, H. J. W. L. Artificial intelligence for clinical oncology. *Cancer Cell* **39**, 916–927 (2021).

9. Chua, I. S. *et al.* Artificial intelligence in oncology: Path to implementation. *Cancer Med.* **10**, 4138–4149 (2021).

10. CONSORT-AI and SPIRIT-AI Steering Group. Reporting guidelines for clinical trials

evaluating artificial intelligence interventions are needed. *Nat. Med.* **25**, 1467–1468 (2019).

11. Leiner, T., Bennink, E., Mol, C. P., Kuijf, H. J. & Veldhuis, W. B. Bringing AI to the clinic: blueprint for a vendor-neutral AI deployment infrastructure. *Insights Imaging* **12**, 11 (2021).

12. Brown, T. B. *et al.* Language Models are Few-Shot Learners. *arXiv* (2020).

13. Johnson, A. E. W. *et al.* MIMIC-CXR-JPG, a large publicly available database of labeled chest radiographs. *arXiv* (2019). https://arxiv.org/abs/1901.07042.

14. Zech, J. R. *et al.* Variable generalization performance of a deep learning model to detect pneumonia in chest radiographs: A cross-sectional study. *PLoS Med.* **15**, e1002683 (2018).

15. Kaushal, A., Altman, R. & Langlotz, C. Geographic distribution of US cohorts used to train deep learning algorithms. *JAMA* **324**, 1212–1213 (2020).

16. Panch, T., Mattie, H. & Celi, L. A. The "inconvenient truth" about AI in healthcare. *npj Digital Med.* **2**, 77 (2019).

17. Winkler, J. K. *et al.* Association between surgical skin markings in dermoscopic images and diagnostic performance of a deep learning convolutional neural network for melanoma recognition. *JAMA Dermatol.* **155**, 1135–1141 (2019).

18. Obermeyer, Z., Powers, B., Vogeli, C. & Mullainathan, S. Dissecting Racial Bias in an Algorithm used to Manage the Health of Populations. *Science* **366**, 447–453 (2019).

19. Castro, D. C., Walker, I. & Glocker, B. Causality matters in medical imaging. *Nat. Commun.* **11**, 3673 (2020).

20. Mattson, P. *et al.* MLPerf Training Benchmark. *arXiv* (2019). https://arxiv.org/abs/1910.01500.

21. Reddi, V. J. *et al.* MLPerf Inference Benchmark. *arXiv* (2019). https://arxiv.org/abs/1911.02549.

22. Reina, G. A. *et al.* OpenFL: An open-source framework for Federated Learning. *arXiv* (2021).

23. Bakas, S. *et al.* Identifying the Best Machine Learning Algorithms for Brain Tumor

Segmentation, Progression Assessment, and Overall Survival Prediction in the BRATS Challenge. *arXiv* (2018). *arXiv* (2018). https://arxiv.org/abs/2105.06413.

24. Pati, S. *et al.* The Federated Tumor Segmentation (FeTS) Challenge. *arXiv* (2021). https://arxiv.org/abs/2105.05874.

25. Placido, D. *et al.* Pancreatic cancer risk predicted from disease trajectories using deep learning. *BioRxiv* (2021) doi:10.1101/2021.06.27.449937.

26. Dayan, I. *et al.* Federated learning for predicting clinical outcomes in patients with COVID-19. *Nat. Med.* (2021) doi:10.1038/s41591-021-01506-3.

27. Larson, D. B., Magnus, D. C., Lungren, M. P., Shah, N. H. & Langlotz, C. P. Ethics of using and sharing clinical imaging data for artificial intelligence: A proposed framework. *Radiology* **295**, 675–682 (2020).

28. Mildenberger, P., Eichelberg, M. & Martin, E. Introduction to the DICOM standard. *Eur. Radiol.* **12**, 920–927 (2002).

29. Sweeney, L. k-Anonymity: A Model for Protecting Privacy. *Int. J. Unc. Fuzz. Knowl. Based Syst.* **10**, 557–570 (2002).

30. Willemink, M. J. *et al.* Preparing medical imaging data for machine learning. *Radiology* **295**, 4–15 (2020).

31. Sambasivan, N. *et al.* "Everyone wants to do the model work, not the data work": Data Cascades in High-Stakes AI. in *Proceedings of the 2021 CHI Conference on Human Factors in Computing Systems* 1–15 (ACM, 2021). doi:10.1145/3411764.3445518.

32. Irvin, J. *et al.* CheXpert: A Large Chest Radiograph Dataset with Uncertainty Labels and Expert Comparison. *AAAI* **33**, 590–597 (2019).

33. Menze, B. H. *et al.* The multimodal brain tumor image segmentation benchmark (BRATS). *IEEE Trans. Med. Imaging* **34**, 1993–2024 (2015).

34. Kramer, B. S., Berg, C. D., Aberle, D. R. & Prorok, P. C. Lung cancer screening with low-dose helical CT: results from the National Lung Screening Trial (NLST). *J. Med. Screen.*

**18**, 109–111 (2011).

35. Kavur, A. E. *et al.* CHAOS Challenge - combined (CT-MR) healthy abdominal organ segmentation. *Med. Image Anal.* **69**, 101950 (2021).

36. Zbontar, J. *et al.* fastMRI: An Open Dataset and Benchmarks for Accelerated MRI. *arXiv* (2018). https://arxiv.org/abs/1811.08839.

37. Hitaj, B., Ateniese, G. & Perez-Cruz, F. Deep Models Under the GAN: Information Leakage from Collaborative Deep Learning. in *Proceedings of the 2017 ACM SIGSAC Conference on Computer and Communications Security - CCS '17* 603–618 (ACM Press, 2017). doi:10.1145/3133956.3134012.

38. Kaissis, G. *et al.* End-to-end privacy preserving deep learning on multi-institutional medical imaging. *Nat. Mach. Intell.* (2021) doi:10.1038/s42256-021-00337-8.

39. Holzinger, A. Interactive machine learning for health informatics: when do we need the human-in-the-loop? *Brain Inform.* **3**, 119–131 (2016).

40. Rieke, N. *et al.* The future of digital health with federated learning. *npj Digital Med.* **3**, 119 (2020).

41. Sheller, M. J. *et al.* Federated learning in medicine: facilitating multi-institutional collaborations without sharing patient data. *Sci. Rep.* **10**, 12598 (2020).

42. Kaissis, G. A., Makowski, M. R., Rückert, D. & Braren, R. F. Secure, privacy-preserving and federated machine learning in medical imaging. *Nat. Mach. Intell.* (2020) doi:10.1038/s42256-020-0186-1.

43. Kairouz, E. by: P. & McMahan, H. B. Advances and open problems in federated learning. *FNT in Machine Learning* **14**, (2021).

44. Girgis, A. M., Data, D., Diggavi, S., Kairouz, P. & Suresh, A. T. Shuffled Model of Federated Learning: Privacy, Communication and Accuracy Trade-offs. *arXiv* (2020). https://arxiv.org/abs/2008.07180.

45. Hernandez, J. B. *et al.* Privacy-first health research with federated learning. *medRxiv* (2020)

doi:10.1101/2020.12.22.20245407.

46. Sirota-Cohen, C., Rosipko, B., Forsberg, D. & Sunshine, J. L. Implementation and Benefits of a Vendor-Neutral Archive and Enterprise-Imaging Management System in an Integrated Delivery Network. *J. Digit. Imaging* **32**, 211–220 (2019).

47. Pantanowitz, L. *et al.* Twenty Years of Digital Pathology: An Overview of the Road Travelled, What is on the Horizon, and the Emergence of Vendor-Neutral Archives. *J. Pathol. Inform.* **9**, 40 (2018).

48. Hripcsak, G. *et al.* Observational Health Data Sciences and Informatics (OHDSI): Opportunities for Observational Researchers. *Stud. Health Technol. Inform.* **216**, 574–578 (2015).

49. Janeway, K.. PRISSMM Data Model. *NCCR Cancer Center Suppl. Data Summit* (2021). https://events.cancer.gov/sites/default/files/assets/dccps/dccps-nccrsummit/08_Katie-Janeway_2021_02_08_PRISSMM.pdf

50. Saripalle, R., Runyan, C. & Russell, M. Using HL7 FHIR to achieve interoperability in patient health record. *J. Biomed. Inform.* **94**, 103188 (2019).

# Contribution

**Author contribution:**

AA: implementation, contribution to concept design, revising the work for intellectual content. AK: co-first author, implementation supervision, contribution to concept design, revising the work for intellectual content, substantial editorial work. AkC: contribution to concept design, revising the work for intellectual content. AlC: contribution to concept design, implementation supervision, revising the work for intellectual content. AR: contribution to concept design, revising the work for intellectual content. AS: contribution to concept design, revising the work for intellectual content. AW: implementation, contribution to concept design, revising the work for intellectual content, substantial editorial work. BD: contribution to concept design, revising the work for intellectual content. CC: contribution to concept design, revising the work for intellectual content. DB: contribution to concept design, revising the work for intellectual content. DD: contribution to concept design, revising the work for intellectual content. DF: contribution to concept design, revising the work for intellectual content. DK: contribution to concept design, revising the work for intellectual content. DX: contribution to concept design, revising the work for intellectual content. GD: contribution to concept design, revising the work for intellectual content. GF: contribution to concept design, revising the work for intellectual content. GP: contribution to concept design, revising the work for intellectual content. IM: contribution to concept design, revising the work for intellectual content. JJ: contribution to concept design, revising work for intellectual content, substantial editorial work. JoG: implementation, contribution to concept design, revising the work for intellectual content. JR: contribution to concept design, revising the work for intellectual content. JT: contribution to concept design, revising the work for intellectual content. JuG: implementation, contribution to concept design, revising the work for intellectual content. MJS: co-first author, implementation supervision, contribution to concept design, revising the work for intellectual content, substantial editorial work. ML: contribution to concept design, revising the work for intellectual content. MR: contribution to concept design, revising the work for intellectual content. MX: contribution to concept design, revising the work for intellectual content. NL: contribution to concept design, revising the work for intellectual content. NN: contribution to concept design, revising the work for intellectual content. NP: contribution to concept design, revising the work for intellectual content. PeM: overall work coordination and supervision, contribution to concept design, revising the work for intellectual content, substantial editorial work. PiM: contribution to concept design, revising the work for intellectual content. PR: contribution to concept design, revising the work for intellectual content. PY: contribution to concept design, revising the work for intellectual content. RU: co-first author, implementation supervision, contribution to concept design, revising the work for intellectual content, substantial editorial work. SB: contribution to concept design, revising the work for intellectual content. SK: contribution to concept design, revising the work for intellectual content. VB: contribution to concept design, revising the work for intellectual content. VM: contribution to concept design, revising the work for intellectual content. VN: contribution to concept design, revising the work for intellectual content. VR: contribution to concept design, revising the work for intellectual content. XH: contribution to concept design, revising the work for intellectual content.

# Conflict of interest

The authors declare that there are no competing interests.